\documentclass{bmvc2k}

\usepackage{amsmath}
\usepackage{booktabs}
\usepackage{subfigure}
\usepackage{array}
\usepackage{wrapfig}
\usepackage{multirow}
\usepackage{tabularx}
\usepackage{amsfonts}
\usepackage{amssymb}
\title{SOFA-Net: Second-Order and First-order Attention Network for Crowd Counting}

\addauthor{Haoran Duan}{h.duan5@newcastle.ac.uk}{1}
\addauthor{Shidong Wang}{shidong.wang@newcastle.ac.uk}{1}
\addauthor{Yu Guan}{yu.guan@newcastle.ac.uk}{1}

\addinstitution{
 Open Lab, School of Computing\\
 Newcastle University\\
 Newcastle Upon Tyne, UK
}

\runninghead{H. DUAN ET AL.}{SOFA-Net for Crowd Counting}

\begin{document}

\maketitle

\begin{abstract}
Automated crowd counting from images/videos has attracted more attention in recent years because of its wide application in smart cities. But modelling the dense crowd heads is challenging and most of the existing works become less reliable. To obtain the appropriate crowd representation, in this work we proposed SOFA-Net(Second-Order and First-order Attention Network): second-order statistics were extracted to retain selectivity of the channel-wise spatial information for dense heads while first-order statistics, which can enhance the feature discrimination for the heads' areas, were used as complementary information. Via a multi-stream architecture, the proposed second/first-order statistics were learned and transformed into attention for robust representation refinement. We evaluated our method on four public datasets and the performance reached state-of-the-art on most of them. Extensive experiments were also conducted to study the components in the proposed SOFA-Net, and the results suggested the high-capability of second/first-order statistics on modelling crowd in challenging scenarios. To the best of our knowledge, we are the first work to explore the second/first-order statistics for crowd counting. 
\end{abstract}

\section{Introduction}
\label{sec:intro}
Crowd counting aims to count the number of people in images or videos of crowd scenes. It plays a pivotal role in real-world applications such as video surveillance, traffic planning, public security, etc. Earlier attempts were based on pedestrian detection \cite{viola2005detecting} or human segmentation \cite{zhao2008segmentation} in crowd. 
Recently, crowd counting has been regarded as an image-based density map regression task, and counting can then be conducted through integrating the densities. 
Density-map based methods achieved promising counting results in crowded scenes when it's difficult to detect subjects due to distance, occlusions, etc.

Previous density map regression works \cite{Idrees_2013_ICCV_Workshops,Pham_2015_ICCV} were proposed to learn the regional objects mapping, and recently Convolution Neural Network (CNN) became the major technique for crowd representation learning. In \cite{sam2017switching}, a switching CNN learning inherent structural and functional differences is proposed to tackle large scale and perspective variations in crowd counting. 
Due to the diverse number of subjects and the various dense or sparse crowd patterns, most recent CNN-based approaches \cite{cao2018scale,Li_2018_CVPR,Liu_2019_CVPR,Liu_2019_ICCV,Xu_2019_ICCV} were proposed to estimate density maps by handling the multi-scale problems in crowd scenes. 
However, these methods become less reliable when the areas of pedestrians' heads are dense and very small. 
In \cite{Cheng_2019_ICCV}, it was found that high density areas tended to be underestimated, while the low density areas tended to be overestimated. 
This observation suggested that a better representation should be learned in such challenging crowd scenarios. So we proposed a deep Second-Order and First-order Attention Network (SOFA-Net) for crowd modelling. Second-order statistics learning was successfully used to improve the representation learning \cite{Li_2017_ICCV,Dai_2017_CVPR,chen2018recurrent,Xia_2019_ICCV} or to recognize small objects in remote sensing \cite{chen2018recurrent,wang2020multi}. In this work, the second-order statistics leads our model to learn robust crowd representation by retaining selectivity of spatial information. First-order statistics, which can capture the discriminated spatial characteristic for crowd, was also used as complementary information.
A Statistic-Wise Convolution operation was also proposed to effectively transform the second/first-order statistics into attentions for our network. 
Then a deep attention architecture was designed to handle multiple feature streams for generating the crowd density maps. 
For better generation quality \cite{zhu2019dual,Liu_2019_CVPR}, a normalization strategy and a scale enhancement were also used.
Our main contribution can be summarized as:

\begin{itemize}
    \item  To the best of our knowledge, this is the first work proposed to use second/first-order statistics for crowd modelling. We analysis the effects of second/first-order statistics for crowd counting qualitatively and quantitatively. Then, the overall experimental results suggested their feasibility in challenging crowd scenarios.
    \item We designed a multi-stream architecture with a Statistic-Wise Convolution to learn the second/first-order statistical attentions for crowd density map generation. Also, several tailored components were also proposed and evaluated.
    \item We tested our method on four popular public datasets, and it reached the state-of-the-art performance on most challenging datasets.
\end{itemize}

\section{Related Work}
\label{sec:related}
\textbf{Crowd Counting}
Early methods were based on designing detection/segmentation algorithms \cite{viola2005detecting,zhao2008segmentation}, yet these methods may be heavily affected by occlusions, making them less practical. In \cite{lempitsky2010learning}, density map estimation approach was first introduced, which aimed to identify the centre locations of the subjects to avoid the error-prone detection procedures. CNN-based methods were the main techniques for representation learning in crowds counting \cite{liu2018crowd,Liu_2019_CVPR,Sindagi_2019_ICCV,zeng2020dspnet}. A maximum-excess-over-pixel loss was proposed with regional feature pattern to utilize the spatial information to count people in different density levels \cite{Cheng_2019_ICCV}. 
In \cite{Liu_2019_ICCV}, Liu et al. proposed a structured feature enhancement module by conditional random fields with a dilated multi-scale structural similarity loss to adapt the scale variations. 

\noindent{\textbf{Second/First Order Statistic} In large scale CNN network, the global average or max pooling was normally set at the end (as the first-order pooling) to capture the image representation by first-order statistical summary\cite{simonyan2014very}. 
Recently, second-order statistics were also explored for improving the representation learning ability in many computer vision tasks \cite{Li_2017_ICCV,Dai_2017_CVPR,chen2018recurrent,Xia_2019_ICCV}. 
Li et.al evaluated the effectiveness of second-order information for large scale visual recognition with a trainable matrix power normalized covariance pooling \cite{Li_2017_ICCV}. The combination of first-order and second-order information was also employed in a multi-level architecture CNN \cite{Dai_2017_CVPR} to improve the image texture discrimination. 
Based on second-order statistics, a recurrent transformer network was proposed \cite{chen2018recurrent} to learn transformation-invariant representation for remote sensing with great performance at recognizing small objects.}

\section{Methodology}
\label{sec:method}
\subsection{Problem Statement}

Following \cite{lempitsky2010learning,Idrees_2013_ICCV_Workshops,Zhang_2016_CVPR,idrees2018composition,Cheng_2019_ICCV}, we describe the crowd counting problem as follows:

Given a crowd image $\mathbf{I}$ with $c$ pedestrians' heads $\mathbf{H} = \{ \mathbf{h}_{i} \in \mathbb{W}^{2} \}^{c}_{i=1}$, where $\mathbf{h}_i$ is the x-y coordinate of the $i_{th}$ center point of the subject's head. 
The (ground truth) density map can be constructed by $c$ Gaussian function $\mathcal{N}$ over all the heads' pixel grids in image $\mathbf{I}$, such that the crowd counts can be calculated by the integrals of density map. The (ground truth) density map can be written as:

\begin{equation}
\mathbf{D}^{gt}=\sum_{\mathbf{h} \in \mathbf{H}} \mathcal{N}(p ; \mu=\mathbf{h}, \sigma^{2})
\end{equation}

\noindent{where $ p \in \mathbf{I}$ denotes the image pixels and $\sigma$ is a very small number (spanning a few pixels \cite{lempitsky2010learning}). Based on $\mathbf{I}$, we can clearly see $c = \sum_{p \in \mathbf{I}} \mathbf{D}^{gt}_p$. 
At the inference stage, given model $\mathcal{F}$ and the query crowd image $\mathbf{I}'$, the density map $\mathbf{D}^{pr}$ and the corresponding counts $c^{pr}$ can be calculated as follows:}

\begin{equation}
\mathbf{D}^{pr} = \mathcal{F}(\mathbf{I}', \mathbf{\hat{W}}), \quad c^{pr} = \sum_{p \in \mathbf{I}} \mathbf{D}^{pr}_p,
\end{equation}

\noindent{where $\mathbf{\hat{W}}$ is the model parameters.}

In previous works, CNN was the major technique for density map regression, yet these models tended to overestimate low density crowd or underestimate high density crowd \cite{Cheng_2019_ICCV}. To learn robust crowd feature, here we propose a deep attention network by exploring the second-order and first-order statistics, and aggregation of these two complementary information may be essential for reliable crowd density estimation. The structure of our method is shown in Fig. 1.

\begin{figure*}[t]
\centering
\includegraphics[width=12cm]{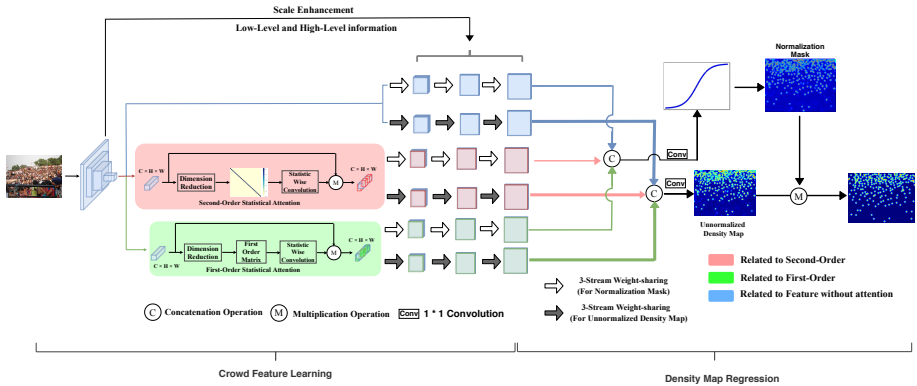}
\caption{The overall framework of SOFA-Net. Pink colored components are related to second-order statistics; Green colored components are related to the first-order statistics, Blue colored components are related to the feature from VGG16 backbone. }
\label{fig:picture001}
\end{figure*}

\subsection{SOFA-Net: Second-Order and First-Order Attention Network}
Fig. 1 shows the overall architecture of our SOFA-Net, which consists of crowd features learning part and the density map regression part. 

The crowd feature learning part starts from a feature encoding backbone, where we use the first 13 layers in VGG16 network \cite{simonyan2014very}. 
Our proposed second/first-order statistical attentions are computed on the VGG16 feature maps at the end of this backbone. Specifically, the second-order statistics can be calculated based on (a derived) covariance matrix, while the first-order statistics can be extracted directly. The second-order and first-order statistics can be learned and transformed into attentions by a proposed Statistic-Wise Convolution operation. 
The two attentions can then be multiplied by crowd feature maps (extracted from VGG16), respectively for the second/first-order based features. 

Given the second/first-order based features as well as the VGG16 features, we can estimate the crowd density map based on the generation blocks containing the Bilinear Up-sampling layer and the basic Convolution operations.
Although the main goal is to learn the second/first-order statistics for crowd representation, we also fuse the VGG16 backbone feature because of the summarized high-level semantic representation \cite{simonyan2014very}. The aggregation of these three feature types may improve the quality of the generated density map. Moreover, a normalization mask is also devised and learned to normalize the density map for better quality. It is worth noting that the normalization mask and the (unnormalized) density map are learned separately, which may make both matrices less correlated for better normalization effect. 
Finally, motivated by the effectiveness of U-Net \cite{ronneberger2015u}, we concatenate the output low-level and high-level crowd features from different backbone layers, which also leads model robust for scale variation. The entire network is optimized by a Pixel-Wise $L_{2}$ loss and a Position-Wise Binary Cross Entropy (BCE) loss \cite{Liu_2019_CVPR}. The details of these components are given the next subsections. 

\subsubsection{Second-Order Statistical Attention}


Recent works \cite{Li_2017_ICCV,chen2018recurrent,Dai_2017_CVPR,Xia_2019_ICCV,wang2020multi} suggested effective representation can be learned by second-order statistics for deep convolution neural network. Here we propose to use second-order statistics as an important component in our SOFA-Net (pink rectangle in Fig. 1), which guides the model to learn the channel-wise spatial information for crowd. Given extracted feature maps (height($h$), width($w$), channel($c$)) from VGG16 backbone $\mathbf{F} \in \mathbb{R}^{h \times w \times c}$, 1$\times$1 convolution, ReLU function and Batch Normalization are applied to reduce the channel dimension from $c$ to $c'$ obtaining $\mathbf{F}' \in \mathbb{R}^{h \times w \times c'}$. Then $\mathbf{F}'$ is flattened into $\mathbf{X} \in \mathbb{R}^{z \times c'}$ where $z = w \times h$. Covariance matrix $\mathbf{C}$ measuring the crowd correlation along channels can be formed as:

\begin{equation}
    \mathbf{C} = \mathbf{X}\mathbf{\overline{I}}\mathbf{X}^T
\end{equation}

\noindent{where $\mathbf{\overline{I}} = \frac{1}{z}(\mathbf{I} - \frac{1}{z}\mathbf{1})$ with $\mathbf{I} \in \mathbb{R}^{c' \times c'}$ the identity matrix and $\mathbf{1} \in \mathbb{R}^{c' \times c'}$ is the matrix of all ones. After reshaping $\mathbf{C}$ to $\mathbf{C}' \in \mathbb{R}^{1 \times c' \times c'}$, the Statistic-Wise Convolution (see Fig. 2) is devised to normalize the covariance to obtain the inherent feature correlation and transform the second-order statistics into attention. The Statistic-Wise Convolution starts from a Independent Statistical Learning (see ISL component in Fig. 2) and the output feature maps $\mathbf{G} \in \mathbb{R}^{1 \times 1 \times m}$ can be formed as:

\begin{equation}
{\mathbf{G}}_{1, 1, m}={\mathbf{K}}_{1 , c', m} \circ \mathbf{C}_{1, c', c'}'
\end{equation}

\noindent{where $\circ$ denotes the element-wise multiplication. The $m(=c')$ convolution kernels $\mathbf{K}$ are applied to the $c'$ channels with the same size $c'\times 1 $ of feature maps (vector in this case).} 
This operation learns the statistical dependency along channels.~Then a 1 $\times$ 1 convolution (i.e., CsL component in Fig. 2) is applied to learn the statistics with sharing convolution kernels and increase the dimension to $c$, so the second-order statistics are transformed into second-order attention as $\mathbf{G} arrow \mathbf{A}_{so}$. This attention $\mathbf{A}_{so}$ is multiplied by the feature $\mathbf{F}$ (extracted from VGG16) to refine the representation.
Specifically, the crowd feature with second-order attention is for
med as $\mathbf{F}_{so} = \mathbf{F} \bigotimes \mathbf{A}_{so}$}, where $\bigotimes$ is the multiplication operation between corresponding feature maps.

\begin{figure}
\centering
\includegraphics[width=9.5cm]{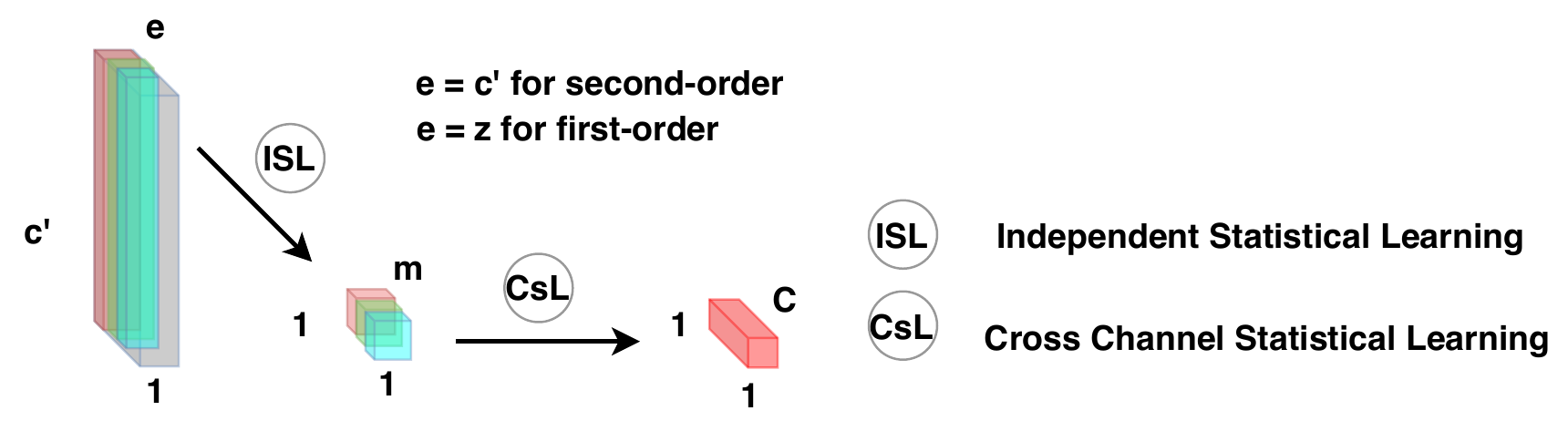}
\caption{Statistic-Wise Convolution including two components: ISL/CsL.}
\label{fig:picture001}
\end{figure}

\subsubsection{First-Order Statistical Attention} 
First-order statistics have been widely adopted in many CNN-based classification tasks to guide the back-propagation \cite{simonyan2014very, qian2020orchestrating}. The first-order statistics were also known to capture the spatial characteristic for texture in images \cite{Dai_2017_CVPR}. 
Here we utilize the first-order statistics to preserve the subjects' head edges with the discrimination of heads and non-heads.
Following similar feature extraction (from VGG16, i.e., with $\mathbf{F} \in \mathbb{R}^{h \times w \times c}$) and dimension reduction procedures (i.e., with $\mathbf{F}' \in \mathbb{R}^{h \times w \times c'}$), we have 
the feature matrix $\mathbf{X} \in \mathbb{R}^{z\times c'}$. 
Different from extracting second-order statistics, we directly use these spatial information.  
Specifically, after reshaping it into $\mathbf{X}_{fo} \in \mathbb{R}^{1 \times z \times c'}$, we learn the first-order attention 
$\mathbf{A}_{fo}$ by performing Statistic-Wise Convolution, i.e.,
${\mathbf{A}_{fo} = \mathbf{K}}_{1 \times z \times c'} \circ \mathbf{X}_{fo}$. 
Similarly, the crowd feature with first-order attention is formed as $\mathbf{F}_{fo} = \mathbf{F} \bigotimes \mathbf{A}_{fo}$. 

\subsubsection{Density Map Estimation}

The crowd density generation process are based on the aforementioned three feature types, i.e., crowd features $\mathbf{F}$ (e.g., extracted from VGG16), features with second-order statistical attention $\mathbf{F}_{so}$, and features with first-order statistical attention $\mathbf{F}_{fo}$.
The final density map $\mathbf{D}^{pr}$ is generated based on two components, namely normalization mask $\mathbf{D}^{msk}$ and unnormalized density map $\mathbf{D}^{udm}$, as shown in Fig. 1. 
To reduce the correlation between these two components for better normalization effect, we train them through two-stream-like structure \cite{zhu2019dual,Liu_2019_CVPR}.

A 3-stream weight-sharing scheme (via generation blocks $f_{G}$ containing Bilinear Up-sampling layer, $1 \times 1$ Convolution layer, $3 \times 3$ Convolution layer, Batch Normalization layer and ReLU activation layer) is used for each component. The unnormalized density map $\mathbf{D}^{udm}$ is generated as:
\begin{equation}
\mathbf{D}^{udm} = \{f_{G}(\mathbf{F},\mathbf{W}^{udm}),f_{G}(\mathbf{F}_{so},\mathbf{W}^{udm}),f_{G}(\mathbf{F}_{fo},\mathbf{W}^{udm})\}.
\end{equation}
Note $\mathbf{W}_{udm}$ are the shared weights among the three feature streams. 
Similarly, the normalization mask can be calculated via $\mathbf{D}^{msk} = f_{sigmoid}(f_{conv}(\mathbf{D'}))$, where
\begin{equation}
\mathbf{D'} = \{f_{G}(\mathbf{F},\mathbf{W}^{msk}),f_{G}(\mathbf{F}_{so},\mathbf{W}^{msk}),f_{G}(\mathbf{F}_{fo},\mathbf{W}^{msk})\},
\end{equation}
and $\mathbf{W}_{msk}$ are the shared weights among the three feature streams.
The final density map can be then estimated through $\mathbf{D}^{pr} =  \mathbf{D}^{udm} \circ \mathbf{D}^{msk}$.

For better training effect \cite{zhu2019dual}, we also applied the scale enhancement strategies, which concatenate the different scale information from different layers in backbone to density map generation block. The scale information allows the model to be more robust for heads' scale variation.

\subsection{Optimization}
To learn the model parameters, we use two loss functions in SOFA-Net: pixel-wise loss and position-wise loss.

\noindent{\textbf{Pixel-Wise Loss} The Pixel-wise loss $L_{2}$ is defined as:

\begin{equation}
L_{2}=\frac{1}{z} \sum_{i=1}^{W} \sum_{j=1}^{H}({\mathbf{D}}_{i, j}^{gt}-\mathbf{D}^{pr}_{i, j})^{2}
\end{equation}

\noindent{where $\mathbf{D}^{pr}$ is the density map with height (H) and width (W) from the feed-forward operation and $z = W \times H$. 
It is the most widely used loss function on training deep convolution networks in crowd counting tasks \cite{Liu_2019_CVPR,Liu_2019_ICCV,Sindagi_2019_ICCV}}.}

\noindent{\textbf{Position-Wise Loss} The position-wise loss $L_{B C E}$ (Binary Cross Entropy) is calculated based on binarized ground truth $\mathbf{D}^{b}$, which comes from the $\mathbf{D}^{gt}$ based on a pre-defined threshold, and the (predicted) normalized density map (via Sigmoid) $\mathbf{S}^{pr} = f_{sigmoid}(\mathbf{D}^{pr})$ as follows:

\begin{equation}
L_{B C E}=-\frac{1}{z} \sum_{p=1}^{z}(\mathbf{D}_{p}^{b} \log (\mathbf{S}^{pr}_{p})+(1-\mathbf{D}_{p}^{b}) \log (1-\mathbf{S}^{pr}_{p}))
\end{equation}

\noindent{The Position-wise loss enforces the learning process to discriminate the crowd locations for better quality of density map generation \cite{Liu_2019_CVPR}}.
\noindent{The final loss function for SOFA-Net optimization is formulated as (in this work, we assign the lambda equal to 0.9).} }

\begin{equation}
L=\lambda L_{2}+(1-\lambda)L_{BCE}
\end{equation}

\begin{table*}[h!]
\centering
\caption{Performance comparison on four public crowd counting datasets}
\scalebox{0.7}{

\begin{tabular}{ |p{2cm}||p{2.3cm}|p{2.4cm}|p{2.3cm}|p{2.3cm}|}
\hline
& UCF\_QNRF & ShanghaiTech A& ShanghaiTech B & UCF\_CC\_50\\
\hline
Method & MAE$\downarrow$\quad MSE$\downarrow$ & MAE$\downarrow$\quad MSE$\downarrow$ & MAE$\downarrow$\quad MSE$\downarrow$ & MAE$\downarrow$\quad MSE$\downarrow$\\
\hline\hline
\cite{Zhang_2016_CVPR}MCNN & \quad - \qquad \quad - &110.2\quad \ 173.2 & \ \ 26.4\quad \ \ 41.3 & 377.6\qquad 509.1 \\
\cite{sam2017switching}SCNN & \quad - \qquad \quad - & \ 90.4\quad \ \ 135.0 & \ \ 21.6\quad \ \ 33.4 & 318.1\qquad 439.2 \\
\cite{Li_2018_CVPR}CSRNet  & \quad - \qquad \quad - & \ 68.2\quad \ \ 115.0 & \ \ 10.6\quad \ \ 16.0 & 266.1\qquad 397.5 \\
\cite{Liu_2019_CVPR}RAZ-Net & \ \  116 \quad  \ \  195.0 & \ 65.1\quad \ \ 106.7 & \ \ 8.40\quad \ \ 14.1 &\quad - \qquad \ \quad -  \\
\cite{Xu_2019_ICCV}L2SM & \  104.7 \quad 173.6 & \ 64.2\quad \ \ 98.40 & \ \ 7.20\quad \ \ 11.1 & 188.4\qquad315.3 \\
\cite{Liu_2019_ICCV}DSSINet & \ \  99.1 \quad \ 159.2 & \ 60.63\quad 96.04 & \ \ 6.85\quad \ \ 10.34 & 216.9\qquad 302.4 \\
\cite{Sindagi_2019_ICCV}MBTTBF & \ \  97.5 \quad \  165.2 & \ 60.2\quad \  \ 94.10 & \ \ 8.00\quad \ \ 15.5 & 233.1\qquad 300.9 \\
\cite{Cheng_2019_ICCV}SPANet & \quad - \qquad \quad - & \ 59.4\quad \ \  92.50 & \ \ \textbf{6.50}\qquad \textbf{9.9} & 232.6\qquad 311.7 \\
\hline
\textbf{Ours} & \ \ \textbf{96.2} \quad \  \textbf{158.7} & \ \textbf{57.5} \quad \ \textbf{92.12} & \ \  6.80\quad \ \ 10.38 & \textbf{185}\qquad \quad \textbf{281} \\
\hline
\end{tabular}
}
\end{table*}

\section{Experiment}
\label{sec:exp}

\subsection{Implementation Details}

\paragraph{Network Setting}
The first 13 VGG16 layers(pre-trained model on ImageNet) were used to initialize the corresponding layers in SOFA-Net. Other parameters were initialized by Gaussian disstribution with zero mean and 0.01 standard deviation. We set batch size to 50 and epoch number to 2000 in our experiments. We performed bilinear interpolation for any images less than 512$\times$ 512, and the images were fed into network after randomly being cropped to 400 $\times$ 400 pixels. Observing that the images were collected from various situations of illumination, we adjusted the images by gamma contrast [0.5, 1.0] with the probability $25\%$. There are a few gray images in some datasets (e.g., ShanghaiTech\_A), so in data augmentation we randomly converted a few (10$\%$) for robust model training.

\noindent{\textbf{Datasets} Our method was evaluated on the four popular public datasets, i.e., UCF\_QNRF \cite{idrees2018composition}, ShanghaiTech \cite{Zhang_2016_CVPR}(Part A and B), and UCF\_CC\_50 \cite{Idrees_2013_ICCV_Workshops}. Out of them, UCF\_QNRF contains large density variations and the subject number ranges from 49 to 12865. UCF\_CC\_50 includes extreme crowd scenes with serious noise. ShanghaiTech part A is very congested with noise, while ShanghaiTech part B is not congested. Following the protocol used in \cite{Liu_2019_CVPR,zhu2019dual}, we generated the ground truth by a fixed Gaussian kernel. Also, the ground truth binary maps were generated by setting the threshold to 0.001 based on the ground truth density maps\cite{Liu_2019_CVPR,zhu2019dual}. For most train/test configurations, we followed the default protocols in the original papers (i.e., UCF\_QNRF \cite{idrees2018composition}, UCF\_CC\_50 \cite{Idrees_2013_ICCV_Workshops}, ShanghaiTech\_AB \cite{Zhang_2016_CVPR}). Notably, due to the limited sample numbers in UCF\_CC\_50, following \cite{Idrees_2013_ICCV_Workshops} we set 5-fold cross validation for evaluation. }

\subsection{Evaluation Metrics}
Following most existing works, the Mean Absolute Error (MAE) and Mean Square Error(MSE) were used as the evaluation metrics.
Given predicted subject number $c^{pr}$ (which can be inferred from $\mathbf{D}^{pr}$, see Eq. (2)), 
for $N$ test crowd images the evaluation metrics can be calculated by:

\begin{footnotesize}
\begin{equation}
MAE=\frac{1}{N}\sum_{i=1}^{N}|c_{i}^{pr}-c_{i}^{gt}|, MSE=\sqrt{\frac{1}{N}\sum_{i=1}^{N}(c^{pr}_{i}-c_{i}^{gt})^{2}}
\end{equation}
\end{footnotesize}

\noindent{where $c^{gt}$ denotes the ground truth heads' counting number.}

\subsection{Experimental Results}

\paragraph{Model Comparison} 
Table 1 shows the results of our SOFA-Net and other state-of-the-arts on four afore-mentioned datasets. 
Our SOFA-Net outperforms others on most of the datasets (except ShanghaiTech Part B), which suggests its effectiveness on general crowd modelling tasks. 
Compared with the most recent works (i.e., methods in 2019) on UCF\_QNRF dataset, SOFA-Net reaches much better results with further error reduction (i.e., in terms of MAE 1.3 - 19.8 and MSE 0.5 - 36.3) than other methods. 
On ShanghaiTech Part A, our method is also much better in terms of both MAE and MSE.
For ShanghaiTech Part B which was collected from shopping street with less crowded scenes, it can be seen that all the methods have good results in this relatively sparse and simple dataset. The performance can be further boosted by fusing other complementary information via the ensemble learning \cite{guan2017ensembles} or multi-stream structure \cite{cheng2019improving,Cheng_2019_ICCV}. Nevertheless, our method outperforms most of the algorithms, and is comparable with state-of-the-art.
UCF\_CC\_50 dataset, which includes very crowded scenes with high-levels of noises, was considered as the most challenging dataset.
We can see our SOFA-Net can model the crowd counting tasks in these extreme conditions well, with much lower errors than other works (i.e., 3.4-234.5 in MAE and 19.9-260.6 in MSE).

\begin{figure}
\centering
\includegraphics[width=8cm]{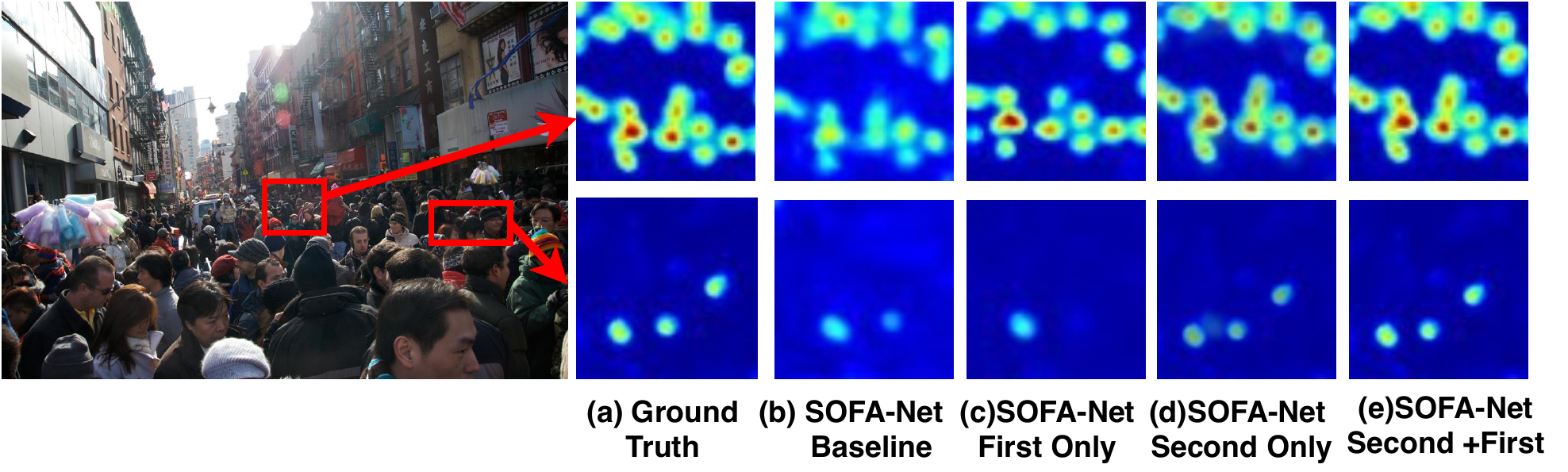}
\caption{The generated maps based on different settings of SOFA-Net in high density area (Top) and low density area (Bottom).}
\label{fig:picture001}
\end{figure}

\begin{figure}
\centering
\includegraphics[width=8cm]{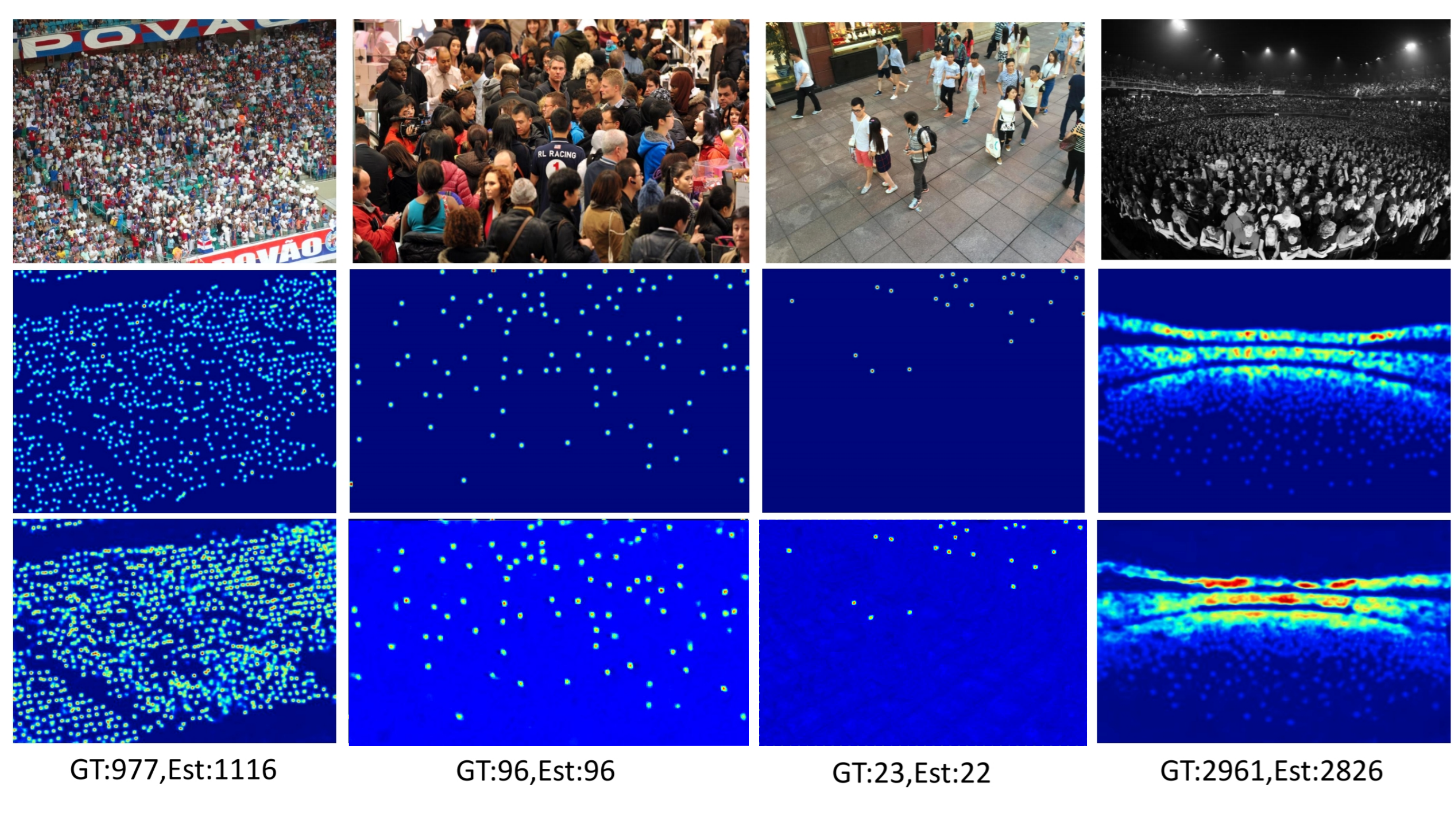}
\caption{Some density maps generated by SOFA-Net; ~From top to bottom: original images, ground truth maps and generated maps}
\label{fig:picture001}
\end{figure}

\noindent{\textbf{Qualitative Analysis} 
To understand better the effect of the proposed second/first-order statistical features, we visualized a challenging crowd image and the generated density maps (under different settings) in Fig. 3, from which some interesting observations can be made: 
\begin{itemize}
	\item from Fig. 3b, we can see without second/first-order statistical features (i.e., with feature $\mathbf{F}$ only), the generated crowd map is very blurry.
	\item with first-order statistical features (Fig. 3c, i.e., with features $\mathbf{F}$,$\mathbf{F}_{fo}$), we can see clear boundaries among heads as enhanced discrimination, yet it cannot model the center-likelihood of dense heads areas well (with relatively low likelihoods in the centers).
	\item Retaining the selectivity of spatial information, features with second-order statistical attention (Fig. 3d, i.e., with features $\mathbf{F}$,$\mathbf{F}_{so}$) can well preserve heads' areas (with high and precise likelihoods), which leads to the accurate counting. Finally aggregating both features can yield precise estimation (Fig. 3e, i.e., with features $\mathbf{F}$,$\mathbf{F}_{fo}$,$\mathbf{F}_{so}$).  
\end{itemize}

From Fig. 3, we can clearly see second/first-order information are complementary for better crowd density map generation. We also generated several density maps in some challenging scenarios (as shown in Fig. 4), and results suggested its effectiveness even when there were more than thousand of people in crowd scenes. }

\subsection{Ablation Study}
We also conducted ablation studies to quantitatively assess the core components in our SOFA-Net.
ShanghaiTech part A, which covers various subject number in different scenes, was used as the benchmark dataset.


\begin{table}[h!]
\centering
\scalebox{0.65}{
\begin{tabular}{ |p{3cm}||p{1.5cm}||p{0.9cm}|p{0.9cm}|  }
 \hline
 \multicolumn{4}{|c|}{ShanghaiTech Part A dataset} \\
 \hline
 SOFA-Net &Features & MAE &MSE\\
 \hline
 No Attention& $\mathbf{F}$	& 68.6	&109.3\\
 First-Order&$\mathbf{F}$ + $\mathbf{F}_{fo}$ &	65.6	& 104.2\\
 Second-Order& $\mathbf{F}$ + $\mathbf{F}_{so}$ &60.8	& 97.1\\
 Second/First-Order &$\mathbf{F}$+$\mathbf{F}_{fo}$+$\mathbf{F}_{so}$ &\textbf{57.5}	& \textbf{92.1}\\
  \hline
 \end{tabular}}
 \caption{on the effect of second/first-order statistical attentions 
 }
\end{table}

\begin{figure}
\centering
\includegraphics[width=12.5cm]{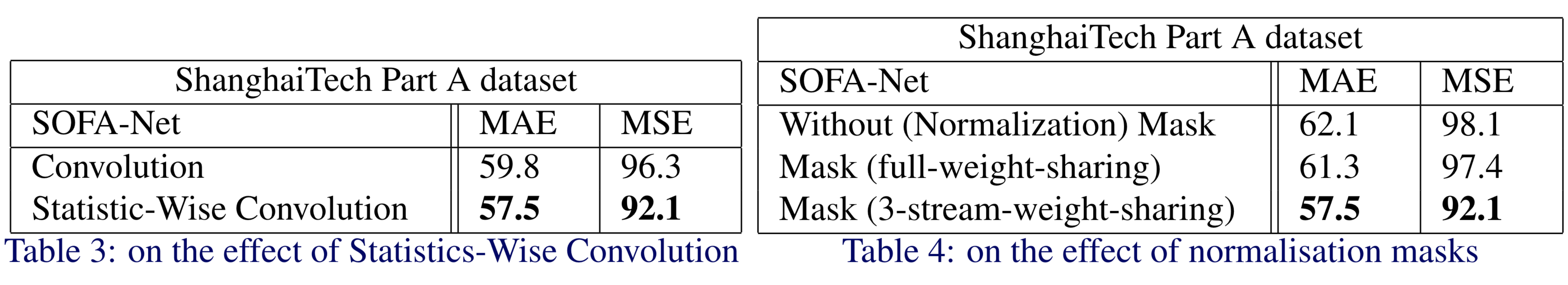}
\label{fig:picture001}
\end{figure}
 
\noindent{\textbf{Effect on Second/First-Order Statistical Attention} The core contribution of this work is the proposed second/first-order statistical attentions for robust representation learning. Fig. 3 demonstrated the effect of both components in a qualitative manner, and here we study them quantitatively. 
In Table 2, we report the SOFA-Net's results under different settings. 
With different attention types, we can use the corresponding feature combinations (e.g., $\mathbf{F}$,$\mathbf{F}_{fo}$,$\mathbf{F}_{so}$) to generate the density maps for crowd counting.  We can clearly see that second-order statistical attention contributes the most to the performance, and the error rate can be further reduced if the complementary second/first-order statistical features were aggregated.}

\noindent{\textbf{Effect on Statistic-Wise Convolution} Statistic-Wise Convolution is a tailored operation that is proposed to learn second/first-order statistical attentions.  The results in Table 3 suggests its effectiveness when compared with the standard convolution operation.}

 
\noindent{\textbf{Effect on Normalization Mask} 
In this work, we also trained a normalization mask to scale the generated density maps to avoid trivial results mostly in non-heads areas. In Table 4, we compared SOFA-Nets with/without normalization masks. For models with normalization masks, we also reported results with two different training strategies, i.e., the proposed 3-stream-weight-sharing scheme, as well as the full-weight-sharing scheme. Specifically, the former shared weights among the three feature streams $\mathbf{F}$,$\mathbf{F}_{fo}$,$\mathbf{F}_{so}$ to learn $\mathbf{W}^{udm}$ in Eq.(5) and $\mathbf{W}^{msk}$ in Eq.(6), respectively, while the latter shared weights among the three feature streams as well as the two tasks (with $\mathbf{W}^{udm}=\mathbf{W}^{msk}$ ). From Table 4, we can clearly see the normalization mask trained by the proposed 3-stream-weight-sharing scheme is much better than other structures. The result also suggests that a less correlated normalization mask (e.g., trained without weight-sharing between tasks) may further reduce the errors in such density map regression tasks.}

\section{Conclusion}
\label{sec:concl}
In this work, we proposed SOFA-Net, which can extract second/first-order statistical attentions to learn robust representations for reliable crowd density map regression. The experimental results suggested second/first-order based features are complementary, and aggregating both features is feasible to reduce the error rate substantially for crowd density estimation. Also, the proposed method outperformed other state-of-the-arts in most (challenging) datasets. Although additional experiments and theoretical findings are necessary to draw the final conclusions on the benefit of applying second/first-order statistics for crowd counting, this work empirically demonstrates a simple yet effective way on modelling the crowd in challenging scenarios.


\bibliography{bmvc_final}
\end{document}